\documentclass{new_tlp}
\usepackage{mathptmx}

\usepackage{times}
\usepackage{graphicx}
\usepackage{latexsym}
\usepackage{algorithm}
\usepackage{float}
\usepackage[noend]{algpseudocode}
\usepackage{amssymb}
\usepackage{amsmath} 
\usepackage{amsfonts}
\usepackage{verbatim} 
\usepackage{longtable}
\usepackage{graphicx}
\usepackage{booktabs}
\usepackage{epstopdf}
\usepackage{longtable}
\usepackage{pgfpages}
\usepackage{multirow}
\usepackage{varwidth}
\usepackage{pifont}
\usepackage{url}

\newtheorem{definition}{Definition}

\newcommand{\myhappens}{\textsf{\footnotesize happensAt}}
\newcommand{\myholdsAt}{\textsf{\footnotesize holdsAt}}
\newcommand{\myinitiatedAt}{\textsf{\footnotesize initiatedAt}}
\newcommand{\myterminatedAt}{\textsf{\footnotesize terminatedAt}}
\newcommand{\mynbf}{\textsf{\footnotesize not}}
\newcommand{\xhail}{\textsf{\footnotesize XHAIL}}
\newcommand{\oled}{\textsf{\footnotesize OLED}}
\newcommand{\iled}{\textsf{\footnotesize ILED}}
\newcommand{\tildee}{\textsf{\footnotesize TILDE}}
\newcommand{\htildee}{\textsf{\footnotesize HTILDE}}
\newcommand{\ec}{\textsf{\footnotesize EC}}
\newcommand{\clingo}{\textsf{\footnotesize Clingo}}

\newcommand{\sq}{\hfill $\square$}

\algnotext{EndFor}
\algnotext{EndIf}

\newenvironment{myexample}{
  \textbf{Example}  
  ~
}{%
\hfill $\square$

}

\title[Online Learning of Event Definitions]
      {Online Learning of Event Definitions}

\author[N. Katzouris, A. Artikis and G. Paliouras]{
NIKOS KATZOURIS$^{1,3}$, ALEXANDER ARTIKIS$^{2,3}$ and GEORGIOS PALIOURAS$^{3}$,  \\
$^1$Department of Informatics \& Telecommunications, National Kapodistrian University of Athens, Athens, Greece\\
$^2$Department of Maritime Studies, University of Piraeus, Piraeus, Greece\\
$^3$Institute of Informatics \& Telecommunications, National Center for Scientific Research ``Demokritos'', Athens, Greece\\
\email{\{nkatz,a.artikis,paliourg\}@iit.demokritos.gr}
}

\begin{document}

\maketitle

\begin{abstract}
Systems for symbolic event recognition infer occurrences of events in time using a set of  event definitions in the form of first-order rules. The Event Calculus is a temporal logic that has been used as a basis in event recognition applications, providing among others,  direct connections to machine learning, via Inductive Logic Programming (ILP). We present an ILP system for online learning of Event Calculus theories. To allow for a single-pass learning strategy, we use the Hoeffding bound for evaluating clauses on a subset of the input stream. We employ a decoupling scheme of the Event Calculus axioms during the learning process, that allows to learn each clause in isolation. Moreover, we use abductive-inductive logic programming techniques to handle unobserved target predicates. We evaluate our approach on an activity recognition application and compare it to a number of batch learning techniques. We obtain results of comparable predicative accuracy with significant speed-ups in training time. We also outperform hand-crafted rules and match the performance of a sound incremental learner that can only operate on noise-free datasets. This paper is under consideration for acceptance in TPLP.  
 
\end{abstract}


\section{Introduction}

Event recognition systems \cite{etzion2010event} process sequences of \emph{simple events}, such as sensor data, and recognize \emph{complex events} of interest, i.e. events that satisfy some pattern. Logic-based event recognition typically uses a knowledge base of first-order rules to represent complex event patterns and a reasoning engine to detect such patterns in the incoming data. Dialects of the Event Calculus (\ec) \cite{kowalski1986logic} have been used as a language for specifying definitions of complex events \cite{artikis2015event}. An advantage of this approach is that is offers direct connections to machine learning, via Inductive Logic Programming (ILP) \cite{de2008logical}, alleviating the task of manual authoring of event definitions. 

Event recognition applications deal with noisy data streams. Methods that extract insights from such streams need to operate within tight memory and time constraints, building a decision model by a single pass over the training data \cite{gama2007learning,gama2010knowledge}. Such a framework is under-explored in ILP, where all data is typically in place when learning begins. Alternatively, some ILP systems are capable of theory revision \cite{esposito2000multistrategy}. Still, such systems need multiple scans of the data to optimize their theories.

We present \oled \ (Online Learning of Event Definitions), an ILP system that learns \ec \  theories in a single pass over a data stream. \oled \ uses the Hoeffding bound \cite{hoeffding1963probability}, a statistical tool that allows to build decision models using only a small subset of the data, by relating the size of this subset to a user-defined confidence level on the error margin of not making a (globally) optimal decision 
\cite{dhurandhar2012distribution,domingos2000mining,gama2011learning}. \oled \ learns a clause in a top-down fashion, by gradually adding literals to its body. Instead of evaluating each candidate specialization on the entire input, it accumulates training data from the stream, until the Hoeffding bound allows to select the best specialization. The instances used to make this decision are not stored or reprocessed, but discarded as soon as \oled \  extracts from them the necessary statistics for clause evaluation.

In the learning problem we address in this work, target clauses are not unrelated, but depend on each other via the axioms of the \ec, making it difficult to use common ILP practices that learn clauses in isolation. To handle this issue we use a decoupling scheme of the axioms of the \ec \  during learning, thereby  allowing to assess the quality of each clause separately, using a scoring function. Additionally, learning programs in the \ec \ involves \emph{non-Observational Predicate Learning} (non-OPL) \cite{muggleton1995inverse}, a setting where instances of the target predicates are not directly observable in the data. To handle non-OPL we use abduction \cite{denecker2002abduction}, a framework that may be used for reasoning with incomplete information. We evaluate our approach on an activity recognition application and compare it to a number of batch learning techniques. We obtain results of comparable predicative accuracy with  significant speed-ups in training time. We also outperform hand-crafted rules and match the performance of a sound incremental learner that can only operate on noise-free datasets.   

The rest of this paper is structured as follows: In Section \ref{sec:related_work} we discuss related work, while in Section \ref{sec:backgound} we present some necessary background on the \ec, ILP and the Hoeffding bound. In  Section \ref{sec:main} we present \oled \ and in Section \ref{sec:experiments} we show the results of the empirical analysis. Finally, in Section \ref{sec:final} we discuss some directions for future research and conclude.

%\vspace*{-0.2cm}

\section{Related work}
\label{sec:related_work}
The Hoeffding bound has been used for propositional machine learning tasks on data streams, such as learning decision trees \cite{domingos2000mining} and  decision rules \cite{gama2011learning},  
and clustering \cite{domingos2001general}. However, its usage for learning relational models is limited. One reason is that it requires independence of observations, which cannot always be ensured in relational domains, due to dependencies in the data \cite{jensen1999statistical,jensen2002autocorrelation,hulten2003mining,dhurandhar2012distribution}.
An ILP approach that uses the Hoeffding bound for relational learning is \htildee \  \cite{lopes2009htilde}, an extension of the \tildee \  system for learning first-order decision trees \cite{blockeel1998top}. These are decision trees where each internal node consists of a conjunction of literals and each leaf is a propositional predicate representing a class. \tildee \  constructs trees by testing conjunctions of literals at each node, using an ILP refinement operator to generate the conjunctions and information gain as the guiding heuristic. \htildee \  extends \tildee \  by using the Hoeffding bound to perform these internal tests on a subset of the training data. To ensure independence of observations, \htildee \  learns from interpretations \cite{blockeel1999scaling}, a setting, used also by \oled, where each training instance is assumed a disconnected part of the dataset. 

Like \tildee, \htildee \ learns clauses with a propositional predicate in the head (representing a class). However, the head of a complex event definition is typically a first-order predicate, containing variables that appear in the body of the clause and express relations between entities. Therefore, \htildee \ is not general enough for the problem we address in this work. Additionally, \htildee \  requires a fully annotated dataset, while in the setting we assume here, annotation for target predicates is missing.

Learning programs in the \ec \ is a challenging task that most ILP learners cannot fully undertake \cite{ray2009nonmonotonic,katzouris2015incremental}, mainly due to the non-monotonicity of negation as failure (NaF) that the \ec \  uses.  \xhail \  \cite{ray2009nonmonotonic} and \textsf{\footnotesize TAL/ASPAL/RASPAL} \cite{athakravi2013learning} are systems that can handle the task, by combining ILP with the non-monotonic semantics of abductive logic programming. These approaches ensure soundness of the outcome, which in the presence of NaF requires learning whole theories by jointly optimizing their clauses. This implies an intractable search space, even with relatively small amounts of data. As a result, the aforementioned approaches do not scale to event recognition applications with temporal data streams. In contrast, \oled \ learns clauses separately using only fragments of the data in an online setting, trading soundness for efficiency. 

\iled \  \cite{katzouris2015incremental} is a recently proposed scalable extension of the \xhail \  system that is able to learn \ec \ theories. It is an incremental learner that revises past hypotheses to fit new observations, and a full-memory system, meaning that revisions should account for a growing historical memory of accumulated data. Using a compressive memory structure to encode the positive examples that each clause entails in the historical memory, \iled \  requires at most one pass over the past data to revise a hypothesis. One difference from \oled \ is that the latter learns in an online fashion, thus it does not re-process past examples. Also, \iled \ is designed to learn sound theories and a key assumption for its scalable strategy is that the training data is noise-free. Other incremental ILP systems, such as INTHELEX \cite{esposito2000multistrategy} and FORTE \cite{richards1995automated}, cannot be applied to the task we address in this work, since they cannot handle negation (FORTE) and non-observable target predicates (INTHELEX, FORTE).

\section{Background and Running Example}  
\label{sec:backgound}   

We assume a logic programming setting, where predicates, terms, atoms, literals, clauses and programs (theories) are defined as in \cite{gebser2012answer} and \textsf{\small{not}} denotes NaF. Following Prolog's convention, predicates and ground terms in logical formulae start with a lower case letter, while variable terms start with a capital letter.

\begin{table}
\footnotesize
\caption{\small The basic predicates and domain-independent axioms of the \ec \  dialect.}
\label{table:ec}
\begin{minipage}{\textwidth}
\begin{tabular}{ll|l}
\hline
\hline
\textbf{Predicate} & \textbf{Predicate Meaning} & \textbf{Axioms} \\
\hline
\myhappens$(E,T)$ & Event $E$ occurs at time $T$ & $\myholdsAt(F,T+1) \leftarrow$\\
\myinitiatedAt$(F,T)$ & At time $T$ a period of time for & $\qquad  \myinitiatedAt(F,T). \  \  \  \  \  \  \  \  \  \  \ (1)$\\
~ & which fluent $F$ holds is initiated & ~ \\
\myterminatedAt$(F,T)$ & At time $T$ a period of time for & $\myholdsAt(F,T+1) \leftarrow $\\
~ & which fluent $F$ holds is terminated & $\qquad \myholdsAt(F,T), \  \  \  \  \  \  \  \  \  \  \  \  \  \  \ (2)$ \\
\myholdsAt$(F,T)$ & Fluent $F$ holds at time $T$ & $\qquad \mynbf \  \myterminatedAt(F,T).$ \\
\hline
\hline
\end{tabular}
\vspace{-2\baselineskip}
\end{minipage}
\end{table}
\normalsize

The Event Calculus (\ec) \cite{kowalski1986logic} is a temporal logic for reasoning about events and their effects. Its ontology comprises time points, represented by integers; fluents, i.e. properties which have certain values in time; and events, i.e. occurrences in time that may affect fluents and alter their value. The axioms of the \ec \  incorporate the \emph{common sense law of inertia}, according to which fluents persist over time, unless they are affected by an event. We use a simplified version of the \ec \ that has been shown to suffice for event recognition \cite{artikis2015event}. The basic predicates and its \emph{domain-independent} axioms are presented in Table \ref{table:ec}. Axiom (1) states that a fluent $F$ holds at time $T$ if it has been initiated at the previous time point, while Axiom (2) states that $F$ continues to hold unless it is terminated.

Definitions of \myinitiatedAt/2 and \myterminatedAt/2 predicates are provided by a set of \emph{domain-specific} axioms. To illustrate our learning approach we use the task of activity recognition, as defined in the CAVIAR project\footnote{\url{http://homepages.inf.ed.ac.uk/rbf/CAVIARDATA1/}}.
The CAVIAR dataset consists of videos of a public space, where actors perform some activities. These videos have been manually annotated by the CAVIAR team to provide the ground truth for two types of activity. The first type, corresponding to simple events, consists of knowledge about a person's activities at a certain video frame/time point (e.g. \emph{walking, standing still} and so on). The second type, corresponding to complex events, consists of activities that involve more than one person, for instance two people \emph{moving together, meeting each other, fighting} and so on. The aim is to recognize complex events by means of combinations of simple events and some additional domain knowledge, such as a person's position and direction.

Table 2(a) presents an example of CAVIAR data, consisting of a narrative of simple events in terms of \myhappens/2, expressing the short-term activities of people, and context properties in terms of \myholdsAt/2, denoting the coordinates and direction of the people. Table 2(a) also shows the annotation of complex events (long-term activities) for each time-point in the narrative. The annotation about complex events is obtained via the closed world assumption (we state both positive and negated annotation atoms in Table \ref{table:stream1} to avoid confusion). An example of two domain-specific axioms in the \ec \  is presented in Table \ref{table:stream1}(b).

\begin{table}
\footnotesize
\caption{\small \textbf{(a)} Example data from activity recognition. For instance, at time point 1 person $\mathit{id_1}$ is \emph{walking}, her $(x,y)$ coordinates are $(201,454)$ and her direction is $270^{\circ}$. The annotation for the same time point states that persons $id_1$ and $id_2$ are not moving together, in contrast to the annotation for time point 2. \textbf{(b)} An example of two domain-specific axioms in the \ec. The first clause dictates that \emph{moving} of two persons $X$ and $Y$ is initiated at time $T$ if both $X$ and $Y$ are walking at time $T$, their euclidean distance is less than $25$ and their difference in direction is less than $45^{\circ}$. The second clause dictates that \emph{moving} of $X$ and $Y$ is terminated at time $T$ if one of them is standing still at time $T$ (exhibits an inactive behavior) and their euclidean distance at $T$ is greater that 30.}
\label{table:stream1}
\begin{minipage}{\textwidth}
\begin{tabular}{lll}
\hline
\hline
\textbf{(a)} & ~ & \textbf{(b)}\\
\textbf{\underline{Narrative for time 1:}} & \textbf{\underline{Narrative for time 2:}} & \textbf{\underline{Two Domain-specific axioms:}}  \\
\noalign{\smallskip}
$\mathit{\myhappens(walking(id_1),1)}$ & $\mathit{\myhappens(walking(id_1),2)}$ & $\myinitiatedAt(moving(X,Y),T) \leftarrow$ \\
$\mathit{\myhappens(walking(id_2),1)}$ & $\mathit{\myhappens(walking(id_2),2)}$ & $\qquad  \myhappens(\mathit{walking(X),T)},$ \\            
$\mathit{\myholdsAt(coords(id_1,201,454),1)}$ & $\mathit{\myholdsAt(coords(id_1,201,454),2)}$ & $\qquad  \myhappens(\mathit{walking(Y),T)},$\\
$\mathit{\myholdsAt(coords(id_2,230,440),1)}$ & $\mathit{\myholdsAt(coords(id_2,227,440),2)}$ & $\qquad  \mathit{distanceLessThan(X,Y,25,T)},$\\
$\mathit{\myholdsAt(direction(id_1,270),1)}$ & $\mathit{\myholdsAt(direction(id_1,275),2)}$ &$\qquad  \mathit{directionLessThan(X,Y,45,T).}$ \\
$\mathit{\myholdsAt(direction(id_2,270),1)}$ & $\mathit{\myholdsAt(direction(id_2,278),2)}$ & ~ \\
\noalign{\smallskip}
\textbf{\underline{Annotation for time 1:}} & \textbf{\underline{Annotation for time 2:}} &$\myterminatedAt(moving(X,Y),T) \leftarrow$ \\
%\noalign{\smallskip}
\mynbf \  \myholdsAt$(moving(id_1,id_2),1)$ & \myholdsAt$(moving(id_1,id_2),2)$ & $\quad  \quad  \myhappens(\mathit{inactive(X),T)},$ \\
~ & ~ & $\quad \quad   \mathit{distanceMoreThan(X,Y,30,T)}.$\\
\noalign{\smallskip}
\hline
\hline
\end{tabular}
\vspace{-2\baselineskip}
\end{minipage}
\end{table}
\normalsize

Our goal is to learn a set of domain-specific axioms specifying complex events. Inductive Logic Programming (ILP) \cite{de2008logical} provides techniques for learning logical theories from examples. In the \emph{Learning from Interpretations} (\emph{LfI}) \cite{blockeel1999scaling} setting that we use in this work, each training example is an interpretation, i.e. a set of true ground atoms, as in Table \ref{table:stream1}(a). Given a set of training interpretations $\mathcal{I}$ and some background theory $B$, which in our case consists of the domain-independent axioms of the \ec, the goal in \emph{LfI} is to find a theory $H$, such that for each interpretation $I\in \mathcal{I}$, $B\cup H$ \emph{covers} $I$, i.e. $I$ is a model of $B\cup H$. Although different semantics are possible, in this work a ``model'' is an \emph{answer set} \cite{gebser2012answer}. 

To allow for an online learning setting, we use the Hoeffding bound \cite{hoeffding1963probability}, a statistical tool that may be used as a probabilistic estimator  of the generalization error of a model (true expected error on the entire input), given its empirical error (observed error on a training subset) \cite{dhurandhar2012distribution}. Given a random variable $X$ with range in $[0,1]$ and an observed mean $\overline{X}$ of its values after $n$ independent observations, the Hoeffding Bound states that, with probability $1 - \delta$, the true mean $\hat{X}$ of the variable lies in an interval $(\overline{X} - \epsilon , \overline{X} + \epsilon)$, where $\epsilon = \sqrt{\frac{ln(1/\delta)}{2n}}$. In other words, the true average can be approximated by the observed one with probability $1-\delta$, given an error margin $\epsilon$ that decreases with the number of observations $n$.

%\vspace*{-0.5cm}

\section{Online Learning of Event Definitions}
\label{sec:main}

ILP learners typically employ a separate-and-conquer strategy: clauses that cover subsets of the examples are constructed one by one recursively, until all examples are covered. Each clause is constructed in a top-down fashion, starting from an overly general clause and gradually specializing it by adding literals to its body. The process is guided by a heuristic function $G$ that assesses the quality of each specialization on the entire training set. At each step, the literal (or set of literals) with the optimal $G$-score is selected and the process continues until certain criteria are met. To adapt this strategy to an online setting, we use the Hoeffding bound to evaluate candidate specializations on a subset of the training interpretations, instead of evaluating them on the entire input. To do so, we use an argument adapted from \cite{domingos2000mining}. Let $r$ be a clause and $G$ a clause evaluation function with range in $[0,1]$. The evaluation function that we use in this work will be discussed shortly. Assume also that after $n$ training instances, $r_1$ is $r$'s specialization with the highest observed mean $G$-score %(the mean $G$-score is denoted by 
$\overline{G}$ and $r_2$ is the second best one, i.e. $\Delta \overline{G} = \overline{G}(r_1)-\overline{G}(r_2) > 0$. Then by the Hoeffding bound we have that for the true mean of the scores' difference $\Delta \hat{G}$ it holds $\Delta\hat{G} > \Delta\overline{G} - \epsilon \text{, with probability } 1 - \delta$, where $\epsilon = \sqrt{\frac{ln(1/\delta)}{2n}}$. Hence, if $\Delta\overline{G} > \epsilon$ then $\Delta \hat{G} > 0$, implying that  $r_1$ is indeed the best specialization to select at this point, with probability $1 - \delta$. In order to decide which specialization to select, it thus suffices to accumulate observations from the input stream until $\Delta\overline{G} > \epsilon$. Since $\epsilon$ decreases with the number of observations, given a desired $\delta$, the number of observations $n$ needed to reach a decision may be traded for a tolerable generalization error $\epsilon$ of not selecting the optimal specialization at a certain choice point. The observations need not be stored or reprocessed. We process each observation once to extract the necessary statistics for the computation of the $G$-score of each candidate specialization. This gives rise to a single-pass clause construction strategy.  

In \emph{LfI} each interpretation is independent form others \cite{blockeel1999scaling}. This guarantees the independence of observations that is necessary for using the Hoeffding bound. In our setting, an interpretation consists of ground atoms $I$ known true at two consecutive time points $T$ and $T{+}1$, as in Table \ref{table:stream1}(a). 
In our \ec \ dialect, the initiation/termination of complex events depends only on the simple events and contextual information of the previous time-point, therefore each interpretation is an independent training instance.

\subsection{Evaluating Clauses}
\label{sec:clause_evaluation}

We relax the \emph{LfI} requirement that a hypothesis $H$ covers every training interpretation to account for noise, and thus seek for a theory with a good fit in the training data. To this end, we define true positive, false positive and false negative atoms as follows:

\begin{definition}[TP, FP, FN atoms]
\label{def:tp_fp_fn}
Let $B$ consist of the domain-independent \ec \ axioms, $r$ be a clause and $I$ an interpretation. We denote by $\mathit{narrative(I)}$ and $\mathit{annotation(I)}$ the narrative and the annotation part of $I$ respectively (see also Table \ref{table:stream1}(a)). We denote by $M^{r}_I$ an answer set of $\mathit{B \cup narrative(I) \cup r}$. Given an annotation atom $\alpha$ we say that: 

\begin{itemize}
\item $\alpha$ is a true positive (\emph{TP}) atom w.r.t. clause $r$ iff $\mathit{\alpha \in annotation(I) \cap M^{r}_I}$.
\item $\alpha$ is a false positive (\emph{FP}) atom w.r.t. clause $r$  iff $\alpha \in M^{r}_I$ but $\mathit{\alpha \notin annotation(I)}$. 
\item $\alpha$ is a false negative (\emph{FN}) atom w.r.t. clause $r$ , iff $\mathit{\alpha \in annotation(I)}$ but $\alpha \notin M^{r}_I$. 
\end{itemize}
\vspace*{-0.4cm}
\sq
\end{definition}

\noindent We seek a theory $H$ that maximizes the \emph{TP} atoms, while minimizing the \emph{FP} and \emph{FN} atoms, collectively for all its clauses. To do so, we maintain a count per clause for each such atom. For an \myinitiatedAt \  clause, its \emph{TP} (resp. \emph{FP}) count  is increased each time it correctly (resp. incorrectly) initiates a complex event (according to the annotation). For a \myterminatedAt \  clause, its \emph{TP} count is increased each time it correctly allows a complex event to persist, by not terminating it. Its \emph{FN} count is increased when it incorrectly terminates a complex event. 

When learning structure in Horn (negation-free) logic with ILP, a theory $H$ is augmented with new clauses to increase its total \emph{TP} count, while existing clauses in $H$ are specialized to decrease the \emph{FP} count. This strategy is not directly applicable to the problem at hand. %, due to the non-monotonicity of the \ec. 
When learning programs in the \ec,  the addition of new clauses may be necessary to eliminate \emph{FPs}, while clause specialization may be necessary to increase \emph{TPs}, as we explain below. Given a theory $H$ and interpretation $I$, assume that $B\cup H$ does not cover $I$. Then one of the following holds: 

\begin{enumerate}
\item \textbf{The \emph{FN} case. } There is at least one \emph{FN} atom $\alpha$. This may be due to one of the following: 

\begin{enumerate}
\item No \myinitiatedAt \  clause in $H$ ``fires'', failing to initiate the complex event that corresponds to $\alpha$, when it should. In this case,  generating a new \myinitiatedAt \ clause, eliminates the \emph{FN} atom, turning it into a \emph{TP}.
\item One or more \myterminatedAt \ clauses in $H$ are over-general, terminating the complex event that corresponds to $\alpha$ when they should not. Specializing the over-general clauses, eliminates the \emph{FN} atom, turning it into a \emph{TP}.
\end{enumerate}

\item \textbf{The \emph{FP} case. } There is at least one \emph{FP} atom $\alpha$. This may be due to one of the following:
 \begin{enumerate}
 \item No \myterminatedAt \  clause in $H$ ``fires'', failing to terminate the complex event that corresponds to $\alpha$ when it should, so $\alpha$ erroneously persists by inertia. Generating a new \myterminatedAt \  clause eliminates the \emph{FP}.
 \item One or more \myinitiatedAt \  clauses are over-general, re-initiating a corresponding complex event when they should not. Specializing the over-general clauses eliminates the \emph{FP}.
 \end{enumerate}  
\end{enumerate}

\noindent Given the different possible behaviours of initiation and termination clauses in the process of optimizing a theory $H$, we next define the clause evaluation function.

\begin{definition}[Clause evaluation function]
\label{def:clause_eval}
Let us denote by $\mathit{TP_{r},FP_{r}}$ and $\mathit{FN_{r}}$ respectively, the accumulated \emph{TP, FP} and \emph{FN} counts of clause $r$ over the input stream. The clause evaluation function $G$ for a clause $r$ is a function with range in $[0,1]$ defined as follows:  

%\begin{equation*}
\ \  \  \  \  \  \  \  \  \  \  \  \  \  \  \  \  \  \  \  \  \  \  \  \  \  \  \  \  \  \ $G(r) = \begin{cases} \frac{TP_r}{TP_r + FP_r} &\mbox{if } r \mbox{ is an } \myinitiatedAt \mbox{ clause }   \\ 
\frac{TP_r}{TP_r + FN_r} &\mbox{if } r \mbox{ is a } \myterminatedAt \mbox{ clause }
  \end{cases}$
%\end{equation*}
\sq
\end{definition}

%\vspace*{-0.3cm}

\noindent Both \myinitiatedAt \ and \myterminatedAt \ clauses affect the total \emph{TP} count of a theory $H$, therefore \emph{TP} counts per clause are taken into account for the evaluation of both types of clauses. Additionally, specializing existing clauses further improves the quality of $H$ by eliminating \emph{FPs} in the \myinitiatedAt \ case (case 2(b) above) and \emph{FNs} in favor of \emph{TPs} in the \myterminatedAt \ case (case 1(b)). Therefore \emph{FPs} (resp. \emph{FNs}) should also be taken into account when evaluating \myinitiatedAt \ (resp. \myterminatedAt) clauses. On the other hand, the total \emph{FP} (resp. \emph{FN}) count of a theory $H$ is not affected by its \emph{existing} \myterminatedAt \ (resp. \myinitiatedAt) clauses, but instead requires new clauses to be generated (cases 2(a) and 1(a) respectively). Therefore \emph{FPs} and \emph{FNs} are irrelevant for the evaluation of existing \myterminatedAt \ and \myinitiatedAt \ clauses respectively. Combining these observations we derive the scoring function of Definition \ref{def:clause_eval}, that uses precision and recall for \myinitiatedAt \ and \myterminatedAt \ clauses respectively. \\

\begin{algorithm}[t]
\small
 \caption{\textsf{\small OnlineLearning}$(\mathcal{I},B,G,\delta,d,\mathit{N_{min}},\mathit{S_{min})}$\newline
 \small
\textbf{Input: }$\mathcal{I}$: A stream of training interpretations; 
$B$: Background knowledge; $G$: Clause evaluation function; $\delta:$ Confidence for the Hoeffding test; $d:$ Specialization depth; 
$S_{min}:$ Clause $G$-score quality threshold.
\normalsize}
 \label{alg:main} 
\begin{algorithmic}[1]
\State $H := \emptyset$ \label{alg:main_start}
\ForAll{$I \in \mathcal{I}$}
  \State \begin{varwidth}[t]{\linewidth} Update $\mathit{TP_{r}, FP_{r}, FN_{r}}$ and $N_r$ counts from $I$, for each $r\in H$ and each $r'\in \rho_d(r)$, \par where $N_r$ denotes the number of examples on which $r$ has been evaluated so far. \end{varwidth}\label{alg:update_statistics}
  \If{$\mathtt{ExpandTheory}(B,H,I)$}\label{alg:expand_theory_test}
    \State $H \leftarrow H \cup \mathtt{StartNewClause}(B,I)$ \label{line:start_new_rule}
  \Else
    %\State $H \leftarrow \mathsf{ExpandExistingClauses}(H)$
    \ForAll{clause $r \in H$}
      \State $r \leftarrow \mathtt{ExpandClause}(r,G,\delta)$
    \EndFor  
  \EndIf
\State $H \leftarrow \mathtt{Prune}(H,S_{min})$  
\EndFor
\State \textbf{return} $H$ \label{alg:main_end}
%~ \\
\State \textbf{function} $\mathtt{StartNewClause}(B,I)$: \label{line:function_start_new_rule}
\State ~ ~ Generate a bottom clause $\bot$ from $I$ and $B$
\State ~ ~ $r := head(\bot) \leftarrow$
\State ~ ~ $\bot_r := \bot$
\State ~ ~ $\mathit{N_r = FP_r = TP_r = FN_r := 0}$ 
\State ~ ~ \textbf{return} $r$ 
\State \textbf{function} $\mathtt{ExpandClause}(r,G,\delta)$: \label{line:function_expand_rule}
\State ~ ~ Compute $\epsilon = \sqrt{\frac{ln(1/\delta)}{2N_r}}$ and let $\overline{G}$ denote the mean value of a clause's $G$-score
\State ~ ~ \begin{varwidth}[t]{\linewidth} Let $r_1$ be the best specialization of $r$, $r_2$ the second best and $\Delta \overline{G} = \overline{G}(r_1)-\overline{G}(r_2)$ \end{varwidth}
\State ~ ~ Let $\tau$ equal the mean value of $\epsilon$ observed so far \label{line:ties}
\State ~ ~ \textbf{if} $\overline{G}(r_1) > \overline{G}(r)$ \textbf{and} [$\Delta \overline{G} > \epsilon$ \textbf{or} $\tau < \epsilon$]: 
\State ~ ~ ~ ~ $\bot_{r_1} := \bot_{r}$
\State ~ ~ ~ ~ \textbf{return} $r_1$
\State ~ ~ \textbf{else} \textbf{return} $r$
\State \textbf{function} $\mathtt{prune}(H, S_{min})$:
\label{line:function_prune}
\State ~ ~ Remove from $H$ each clause $r$ for which $S_{min} - \overline{G}(r) > \epsilon$, where $\epsilon$ is the current Hoeffding bound
\State ~ ~ \textbf{return} $H$
\end{algorithmic}
\end{algorithm}
\normalsize

%\vspace*{-1cm}

\subsection{The \textsf{OLED} system}
In this section we discuss the main functionality of \oled, presented in Algorithm \ref{alg:main}, in detail. Learning begins with an empty hypothesis $H$. On the arrival of new interpretations, \oled  \ either expands $H$, by generating a new clause, or tries to expand (specialize) an existing clause. Clauses of low quality are pruned, after they have been evaluated on a sufficient number of examples. Each incoming interpretation is processed once, to extract the necessary statistics for clause evaluation in the form of \emph{TP}, \emph{FP} and \emph{FN} counts, and is subsequently discarded.

\begin{table}
\footnotesize
\caption{\small Action dispatching scheme for \textsf{OLED}'s  \myinitiatedAt \ ($\mathit{L_{\mathsf{init}}}$) and \myterminatedAt \ ($\mathit{L_{\mathsf{term}}}$) parallel processes. The justification refers to the different cases analysed in Section \ref{sec:clause_evaluation}}\label{table:control}
\begin{minipage}{\textwidth}
\begin{tabular}{cccc}
\hline
\hline
\textbf{Process} & \textbf{Cause of Failure} & \textbf{Action} & \textbf{Justification}  \\
\noalign{\smallskip}
$\mathit{L_{\mathsf{init}}}$ & \emph{FP} & Clause expansion  & Case 2(b)\\

$\mathit{L_{\mathsf{init}}}$ & \emph{FN} & Theory expansion & Case 1(a)\\ 

$\mathit{L_{\mathsf{term}}}$ & \emph{FP} & Theory expansion & Case 2(a)\\

$\mathit{L_{\mathsf{term}}}$ & \emph{FN} & Clause expansion & Case 1(b)\\
\hline
\hline
\end{tabular}
\vspace{-2\baselineskip}
\end{minipage}
\end{table}
\normalsize

To distinguish between the different cases presented in Section \ref{sec:clause_evaluation}, initiation and termination clauses are learnt separately in parallel, by two processes $\mathit{L_{\mathsf{init}}}$ and $\mathit{L_{\mathsf{term}}}$ respectively (each one of these processes runs separately Algorithm \ref{alg:main}). The input stream is forwarded to each of these processes simultaneously. Thanks to this decoupling, when either process fails to account for a training interpretation, it is able to infer the causes of failure in terms of \emph{FP} and \emph{FN} atoms. In particular $\mathit{L_{\mathsf{init}}}$ detects \emph{FP}/\emph{FN}-failures based on cases 2(b)/1(a) respectively and $\mathit{L_{\mathsf{term}}}$ detects \emph{FP}/\emph{FN}-failures based on cases 2(a)/1(b). According to the cause of failure, the process dispatches control either to the theory expansion, or the clause expansion subroutines. The choice among these actions is made by the boolean function \texttt{ExpandTheory} in line \ref{alg:expand_theory_test} of Algorithm \ref{alg:main}. Action selection is based on the analysis of Section \ref{sec:clause_evaluation} and summarised in Table \ref{table:control}. Below we present an example for illustration purposes.%provides a simple illustration of how the two processes and the action dispatching scheme from Table \ref{table:control} work.

\begin{myexample}
\label{example1}
Initially, processes $\mathit{L_{\mathsf{init}}}$ and $\mathit{L_{\mathsf{term}}}$ start with two empty hypotheses, $\mathit{H_{\mathsf{init}}}$ and $\mathit{H_{\mathsf{term}}}$. %, representing the \myinitiatedAt \ and \myterminatedAt \ parts of the target hypothesis respectively. 
Assume that the annotation in one of the incoming interpretations dictates that the \emph{moving} complex event holds at time $10$, while it does not hold at time $9$. Since no clause in $\mathit{H_{\mathsf{init}}}$ yet exists to initiate \emph{moving} at time $9$, %so that the observation at time $10$ be explained, 
$\mathit{L_{\mathsf{init}}}$ detects the \emph{moving} instance at time $10$ as an \emph{FN} and proceeds to theory expansion (second case in Table \ref{table:control}), generating an initiation clause for \emph{moving}. $\mathit{L_{\mathsf{term}}}$ is not concerned with initiation conditions, so it will take no actions in this case. Then, a new interpretation arrives, where the annotation dictates that \emph{moving} holds at time $20$, but does not hold at time $21$. In this case, since no clause yet exists in $\mathit{H_{\mathsf{term}}}$ to terminate \emph{moving} at time $20$, %it will persist by inertia, and 
$\mathit{L_{\mathsf{term}}}$ will detect an \emph{FP} instance at time $21$. It will then proceed to theory expansion (third case in Table \ref{table:control}), generating a new termination condition for moving. At the same time, assume that the initiation clause in $\mathit{H_{\mathsf{init}}}$ is over-general  and erroneously re-initiates \emph{moving} at time $20$, generating an $FP$ instance for the $\mathit{L_{\mathsf{init}}}$ process at time $21$. In response to that, $\mathit{L_{\mathsf{init}}}$ will proceed to clause expansion (first case in Table \ref{table:control}), penalizing the over-general initiation clause by increasing its \emph{FP} count, thus contributing towards its potential replacement by one of its specializations.
\end{myexample}

In the remainder of this section, we go into the details of theory and clause expansion, as well as some other aspects of \oled's functionality.

\textbf{Theory Expansion. }
The theory expansion process is handled by the \texttt{StartNewClause} function in Algorithm \ref{alg:main}. A new clause is generated in a data-driven fashion, by constructing a \emph{bottom clause} $\bot$ from a training interpretation \cite{muggleton1995inverse}. Theory expansion consists of the addition of the  clause $r = head(\bot) \leftarrow$ to theory $H$. From that point on, $r$ is gradually specialized by the addition of literals from $\bot$ to its body. We denote by $\bot_r$ the bottom clause associated to clause $r$.

In a typical ILP setting, a bottom clause is constructed by selecting a target predicate instance $e$ as a ``seed'', placing it in the head of a newly generated clause $\bot$ with an empty body. A set of atoms that follow deductively from $e$ and the background knowledge are placed in the body of $\bot$. Constants in $\bot$ are replaced with variables, where appropriate, as indicated by a particular language bias, typically \emph{mode declarations} \cite{muggleton1995inverse}. To find a clause with a good fit in the data, a \emph{refinement operator} $\rho$ is used to generate candidate clauses that $\theta$-subsume $\bot$. 

The aforementioned approach cannot be applied directly to the problem we address here, which falls in the non-Observational Predicate Learning (OPL) class of problems \cite{muggleton1995inverse}. In non-OPL, instances of target predicates, that are normally used as seeds for the construction of $\bot$, are not directly observable in the training data. In our case, target predicates are \myinitiatedAt/2 and \myterminatedAt/2, while the annotation in the training interpretations consists of complex event instances in terms of the \myholdsAt/2 \ predicate (see Table \ref{table:stream1}). A workaround is to use abduction %\cite{denecker2002abduction} 
to obtain the missing target predicate instances and then construct bottom clauses from them. This approach is followed by the \xhail \ system \cite{ray2009nonmonotonic} and we also adopt it here. Like \xhail, \oled \  also uses mode declarations for specifying the language bias. 

\oled \ %is an any-time algorithm, i.e. it 
may output the hypothesis constructed so far at any time during the learning process. We allow a ``warm-up'' period, in the form of a minimum number of training instances $\mathit{N_{min}}$ on which a clause $r$ must be evaluated before it can be included in an output hypothesis.

\textbf{Clause Expansion.} %\oled's clause expansion process has been introduced earlier in this section. 
We use the Hoeffding bound to select among competing specializations of a clause $r$. These specializations are generated by adding one or more literals from $\bot_r$ to the body of $r$. An input parameter $d$ for \emph{specialization depth} serves as an upper bound to the number of literals that may be added at each time. We use $\rho_d(r)$ to denote the set of specializations for clause $r$. Formally, $\rho_d(r) = \{head(r) \leftarrow body(r) \wedge D \  | \  D \subset body(\bot_r) \text{ and } |D| \leq d\}$. E.g. $\rho_1(r)$ consists of all ``one-step'' specializations of $r$ (i.e. those that result by the addition of a single literal from $\bot_r$), while $\rho_2(r)$ consists of $\rho_1(r)$ plus all ``two-step'' specializations, and so on.

A clause $r$ is expanded, i.e. replaced by its best-scoring specialization from $\rho_d(r)$, when a sufficient number of interpretations have been seen, such that $\Delta \overline{G} > \epsilon$, %as described in Section \ref{sec:main}, 
where $\Delta \overline{G}$ is the observed difference between the mean $G$-scores of $r$'s best and second best specializations. %where $\Delta \overline{G}$ is the observed difference between the mean $G$-scores of $r$'s best and second best specializations and $\epsilon$ is the current Hoeffding bound. %To compute the running mean difference for each clause across the stream, all that is required is an extra variable (per clause) that stores the previous observed value. 
To ensure that no clause $r$ is replaced by a specialization of lower quality, $r$ itself is also considered as a potential candidate along with its specializations from $\rho_d(r)$. This ensures that expanding a clause to its best-scoring specialization is better, with probability $1-\delta$, than not expanding it at all.

Online %and incremental 
learners are typically subject to order effects, i.e. they are sensitive to the order in which the examples are presented. Using the Hoeffding bound allows \oled \ to mitigate such effects, since clause expansion is postponed until sufficient evidence for the quality of the candidate specializations is provided by the data. 

\textbf{Tie-breaking. } When the scores of two or more specializations are very similar, a large number of training instances may be required to decide between them. This could be wasteful, since any one of the %equally good 
specializations may be chosen. In such cases, as in \cite{domingos2000mining}, we break ties as follows: Instead of waiting until $\Delta \overline{G} > \epsilon$, as required by the Hoeffding bound-based heuristic, %(see also Section \ref{sec:main})
we expand $r$ to its best-scoring specialization if $\Delta \overline{G} < \epsilon < \tau$, where $\tau$ is a tie-breaking threshold (recall that $\epsilon$ decreases with the number of training examples, thus it may fall below $\tau$). We follow \cite{yang2011moderated} and use an adaptive tie-breaking threshold, set to the mean value of $\epsilon$ that has been observed so far in the training process (see line \ref{line:ties}, Algorithm \ref{alg:main}). In the case of a tie between $r$ itself and its best-scoring specialization, we follow a conservative approach and do not expand $r$.

\textbf{Clause pruning. } 
\oled \ supports removal of clauses whose score is smaller than a quality threshold $\mathit{S_{min}}$. To decide when a clause may be removed we also use the Hoeffding bound. If $S_{min} - \overline{G}(r) > \epsilon$, %where $\epsilon$ is the current Hoeffding bound, 
then with probability $1 - \delta$, the true mean of $r$'s $G$-score is lower than the quality threshold $S_{min}$ and therefore $r$ should be removed.

\begin{table}
\footnotesize
\caption{Experimental results from the CAVIAR dataset}\label{table:results}
\begin{minipage}{\textwidth}
\begin{tabular}{cccccccc}
\hline
\hline
~ & ~ & \textbf{Method} & 
\textbf{Precision} &
\textbf{Recall} & $\mathbf{F_1}$\textbf{-score} & \textbf{Theory size} & \textbf{Time (sec)} \\
\noalign{\smallskip}
\noalign{\smallskip}
\hspace*{-2cm}(a)%\footnote{Results using the CAVIAR fragment from \cite{skarlatidis2015probabilistic}, see Section \ref{sec:experiment1}.}

~ & \emph{Moving} & $\mathsf{EC_{crisp}}$  & \textbf{0.909} & 0.634 & 0.751 & 28  & -- \\
 & ~ & $\mathsf{EC_{MM}}$ &  0.844 & 0.941 & \textbf{0.890} & 28  & 1692\\
~ & ~ & \xhail  &  0.779 & 0.914 & 0.841 & \textbf{14}  & 7836   \\
~ & ~ & $\mathsf{OLED}$ &  0.709 & \textbf{0.948} & 0.812 & 34  & \textbf{12}\\

\noalign{\smallskip}
~ & \emph{Meeting} & $\mathsf{EC_{crisp}}$ & 0.687 & 0.855 & 0.762 & 23  & --\\
~ & ~ & $\mathsf{EC_{MM}}$ & 0.919 & 0.813 & \textbf{0.863} & 23  & 1133\\
~ & ~ & \xhail &  0.804 & \textbf{0.927}  & 0.861 & \textbf{15}  & 7248  \\
~ & ~ & $\mathsf{OLED}$ &  \textbf{0.943} & 0.750 & 0.836 & 29  & \textbf{23 }\\

\hline\noalign{\smallskip}
\noalign{\smallskip}
\hspace*{-2cm}(b) & \emph{Moving} & $\mathsf{EC_{crisp}}$ &  \textbf{0.721} & 0.639 & 0.677 & 28 & -- \\
~ & ~ & $\mathsf{OLED}$ & 0.653 & \textbf{0.834} & \textbf{0.732} & 42 & 124 \\
\noalign{\smallskip}
~ & ~ & $\mathsf{EC_{crisp}}$ & 0.644 & 0.855 & 0.735 & 23 & --\\
~ & \emph{Meeting} & $\mathsf{OLED}$ & \textbf{0.678} & \textbf{0.953} & \textbf{0.792} & 30  & 107\\

\hline\noalign{\smallskip}
\hspace*{-2cm}(c) & \emph{Moving} & $\mathsf{ILED}$  & 0.947 & \textbf{0.981} & \textbf{0.963} & 55  & \textbf{34 } \\
~ & ~ & $\mathsf{OLED}$ &  \textbf{0.963} & 0.934 & 0.948 & \textbf{31}  & 35 \\
%~ & \emph{Meet} & ~ & ~ & ~ & ~ & ~ & ~ \\
~ & \emph{Meeting} & $\mathsf{ILED}$ & 0.930 & \textbf{0.976} & 0.952 & 65  & \textbf{30 }\\
~ & ~ & $\mathsf{OLED}$ &  \textbf{0.975} & 0.933 & \textbf{0.953} & \textbf{53}  & 42 \\
\hline
\hline
\end{tabular}
\vspace{-2\baselineskip}
\end{minipage}
\end{table}
\normalsize

%\vspace*{-0.2cm}

\section{Experimental Evaluation}
\label{sec:experiments}

We evaluate \oled's performance on CAVIAR (see Section \ref{sec:backgound}), a benchmark dataset for activity recognition. CAVIAR contains a total of 282067 training interpretations with a mean size of 25 atoms each. The size of the search space (clause subsumption lattice) %during the learning process  
is determined by the size of bottom clauses, which in these experiments consisted on average of 15 literals each.   %An average bottom clause constructed during the learning process consists of approximately 15 literals, which implies that the number of nodes in the subsumption lattice of a candidate clause (search space complexity) is approximately $2^{15}$.    
All experiments were conducted on a Linux machine with a 3.6GHz processor (4 cores and 8 threads) and 16GiB of RAM. The code and data are available online\footnote{\url{https://github.com/nkatzz/OLED}}.

%\vspace*{-0.4cm}

\subsection{Comparison with Manually Constructed Rules and Batch Learning}
\label{sec:experiment1}
The purpose of this experiment was to assess whether \oled \ is able to efficiently learn theories of comparable quality to hand-crafted rules and state-of-the-art batch learning approaches. We compare \oled \ to the following: (i) $\mathsf{EC_{crisp}}$, a hand-crafted set of clauses for the CAVIAR dataset, described in \cite{artikis2010behaviour}; (ii) $\mathsf{EC_{MM}}$ \cite{skarlatidis2015probabilistic}, a probabilistic version of $\mathsf{EC_{crisp}}$ with weights learnt by the Max-Margin weight learning method for Markov Logic Networks (MLNs) of \cite{huynh2009max}; (iii) \xhail \  \cite{ray2009nonmonotonic}, a hybrid abductive-inductive learner capable of learning programs in the \ec. $\mathsf{\small EC_{MM}}$ was selected because it was shown to achieve good results on CAVIAR \cite{skarlatidis2015probabilistic}. \xhail \ was selected as one of the few ILP systems that is able to learn theories in the \ec.  \oled \  and \xhail \ were implemented using the \clingo\footnote{\url{http://potassco.sourceforge.net/}} answer set solver as the core reasoning component, while the $\mathsf{\small EC_{MM}}$ approach used in this experiment  was implemented in the \textsf{\small LoMRF} framework\footnote{\url{https://github.com/anskarl/LoMRF}} for MLNs.

To evaluate $\mathsf{EC_{MM}}$, \cite{skarlatidis2015probabilistic} used a fragment of the CAVIAR dataset, which is also the one we use in this experiment. The target complex events in this dataset are related to two persons \emph{meeting each other} or \emph{moving together} and the training data consists of the parts of CAVIAR that involve these complex events. The fragment dataset contains a total of 25738 training interpretations. There are 6272 interpretations in which \emph{moving} occurs and 3722 in which \emph{meeting} occurs. \oled's results were achieved using significance $\delta = 10^{-5}$, a clause pruning threshold $S_{min}$ of $0.7$ for \emph{meeting} and $0.5$ for \emph{moving} and specialization depth parameter $d=2$ for \emph{meeting} and $d=1$ for \emph{moving}. The results reported with this parameter configuration are the best among several other parameter settings that we tried for $S_{min}$ and $d$. The training time for each run of \oled \  was the maximum training time of the two parallel processes $\mathit{L_{\mathsf{init}}}$ and $\mathit{L_{\mathsf{term}}}$. 

Results were obtained using 10-fold cross validation and are presented in Table \ref{table:results}(a) in the form of \emph{precision, recall} and $f_1$-\emph{score}. These statistics were micro-averaged over the instances of recognized complex events from each fold of the 10-fold cross validation process. Table \ref{table:results}(a) also presents average training time per fold for all approaches except $\mathsf{EC_{crisp}}$ (where no training is involved), average theory sizes (total number of literals) for \oled \  and \xhail, as well as the fixed theory size of $\mathsf{EC_{crisp}}$ and $\mathsf{EC_{MM}}$.   

$\mathsf{\small EC_{MM}}$ achieves the best $f_1$-score for both complex events, followed closely by \xhail. \oled \  achieves a comparable predictive accuracy (particularly for \emph{meeting}), while it outscores the hand-crafted rules. Moreover, \oled \  achieves speed-ups of several orders of magnitude as compared to $\mathsf{\small EC_{MM}}$ and \xhail, due to its single-pass strategy. The superior accuracy of $\mathsf{EC_{MM}}$ and \xhail \  is due to them being batch learners, optimizing their respective outcomes over the entire training set. This also explains the increased training times for both. %, in addition to the fact that their main operations are computationally expensive. 
Regarding theory size, \xhail \ learns significantly more compressed hypotheses than \oled. The reason is that \xhail \ learns whole theories, while \oled \ learns each clause separately to gain in efficiency.

%\vspace*{-0.3cm}

\subsection{Activity Recognition on the Entire CAVIAR Dataset}
\label{sec:experiment2}

We also present experimental results from running \oled \ on the entire CAVIAR dataset. %, which contains a total of 282067 training interpretations. 
The target complex events are  \emph{meeting} and \emph{moving} as previously. The number of positive interpretations for both complex events is also the same as before, since the data fragment used in the previous experiment contains the parts of CAVIAR where these complex events occur. In contrast, the number of negative training instances is much larger in this experiment.

Due to the high training times of \xhail \ and $\mathsf{EC_{MM}}$, we do not present results with these approaches, and compare \oled \ only to the set of manually developed clauses $\mathsf{EC_{crisp}}$. The experimental setting was as follows: We used 10-fold cross validation over the fragment used in the previous experiment, but in each fold, the training and test sets were augmented by a number of negative training sequences. In particular, in each fold, 90\% of the negative training sequences from the remaining part of CAVIAR (i.e. the part not contained in the data fragment of the previous experiment) was added to the training set of the fold, while the remaining 10\% was added to the test set. The parameter configuration for \oled \  was the same as in the previous experiment, with the exception of the specialization depth for \emph{meeting}, which was set to $d=1$. The limited size of the training sets in the experiment of Section \ref{sec:experiment1} prevented \oled \ from sufficiently expanding its clauses when $d=1$, resulting in over-general theories. Setting $d=2$, thus trying 2-step specializations as well, made it possible to obtain the results reported in Table \ref{table:results}(a). In contrast, this was not necessary in this experiment, where due to the significantly larger training set size, \oled \ was able to find good clauses by trying 1-step specializations only.

Table \ref{table:results}(b) shows the results. 
%As previously, statistics were micro-averaged over the recognized complex event instances from each fold. 
Both approaches' performance is decreased, as compared to the previous experiment, due to the increased number of false positives, caused by the large number of additional negative instances. \oled \ still outscores the hand-crafted knowledge base.

%\vspace{-0.5cm}

\subsection{Comparison with an Incremental Learner}
\label{sec:experiment3}

We compared \oled \ to \iled \ \cite{katzouris2015incremental}, an incremental learner that  is able to learn theories in the \ec. Recall that \iled \ cannot learn from noisy data (see also Section \ref{sec:related_work}), therefore, it cannot be used in CAVIAR, which exhibits various types of noise -- see \cite{artikis2010behaviour} for details. In order to compare the two systems, we thus generated a noise-free version of CAVIAR with artificial annotation for the \emph{moving} and \emph{meeting} complex events. To produce the annotation, we used the hand-crafted knowledge base $\mathsf{\small EC_{crisp}}$ for inference over the CAVIAR narrative. %The dataset contains a total of 282067 training interpretations (6172 are positive interpretations for \emph{meeting} and 5204 are positive interpretations for \emph{moving}).
We used 10-fold cross validation to assess the performance of the the compared systems. For each fold, the training (resp. test) set consisted of the 90\% (resp. 10\%) of positive and negative interpretations for each complex event. \oled's parameter setting was as reported in Section \ref{sec:experiment2}.

The results are presented in Table \ref{table:results}(c). Predictive accuracy for both systems is comparable, with \iled's being slightly better. This was expected, since \iled \ re-scans the historical memory of past data to revise its theories. Training times are also comparable, with \oled's being slightly higher, as compared to \iled's. \iled \ is able to avoid certain computations by inferring that they are redundant, based on the assumption that the data is noise-free. Regarding theory size, \oled \  learns significantly shorter hypotheses that \iled. \oled \ prunes a number of its learnt clauses, in an effort to avoid fitting potential noise in the data and also follows a conservative clause expansion strategy. In contrast, \iled \ tries to account for every positive example (and exclude every negative one), since it is designed for learning sound hypotheses.

%\vspace{-0.3cm}

\subsection{Scalability}
\label{sec:experiment4}

\begin{figure}[t]
\centering
\includegraphics[width=0.6\textwidth]{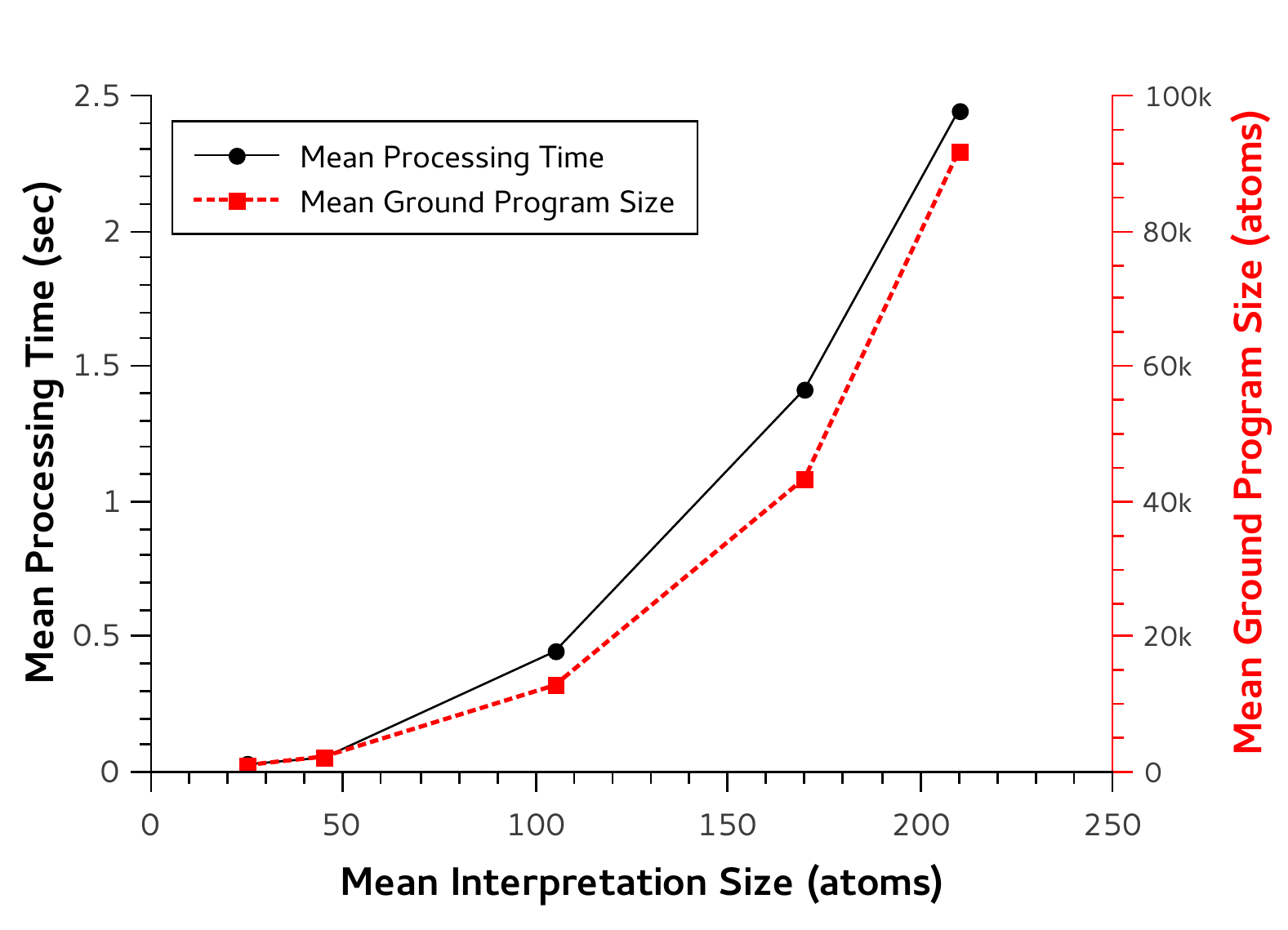}
\caption{\small \oled's mean processing time and mean ground program size per training interpretation, for varying interpretation sizes.}
\label{fig:scalability}
\end{figure}

In this experiment we assess \oled's scalability. When learning from the entire CAVIAR dataset (Section \ref{sec:experiment2}) the average processing time per training interpretation was 6.7 milliseconds (ms), while the frame rate in CAVIAR, i.e. the rate in which video frames containing new data arrive is 40 ms. As a ``stress-test'', we evaluated \oled's performance in more demanding learning tasks. We generated 4 different datasets, each of which consisted of a number of copies of CAVIAR. The new datasets differ from the original one in the constants referring to the tracked entities in simple and complex events. We generated datasets consisting of 2, 5, 8 and 10 copies, each of which contained 20, 50, 80 and 100 different entities respectively. 
Like in the previous experiments, each interpretation includes narrative and annotation atoms from two time points. In this experiment however, the number of atoms in each interpretation  grows proportionally to the number of copies of the dataset. 

We performed learning with \oled \ on the original and the enlarged datasets and measured the average processing time per training interpretation. Figure \ref{fig:scalability} presents the results. For instance, interpretations in the 10 copies of CAVIAR are handled in approximately 2.5 sec in a standard desktop computer. The growth in average processing time is due to the increased number of annotation atoms in the datasets, as well as the additional domain constants, that result in an exponential increase in the size of the ground program produced during the clause evaluation process (see the dashed line in Figure \ref{fig:scalability}). \oled's performance may be improved by some optimizations, such as taking advantage of domain knowledge about relational dependencies in the data. For instance, in CAVIAR complex events involve two different entities, therefore learning may be split across different processing cores that learn from independent parts of the data. Such optimizations are part of our current work.

%\vspace*{-0.4cm}    

\section{Conclusions and Further Work}
\label{sec:final}

We presented \oled, an ILP system for online learning of complex event definitions in the Event Calculus. \oled  \ is an any-time system that learns by a single-pass over a stream, using the Hoeffding bound to evaluate candidate clauses on a subset of the input. Results of the empirical evaluation indicate that \oled \  achieves speed-ups of several orders of magnitude, as compared to batch learners, with a comparable predictive accuracy. It also outscores hand-crafted rules and matches the performance of a sound incremental learner that can only operate on noise-free datasets. We intend to improve \oled \ in several aspects, including scalability and development of adaptive techniques for automated configuration of its parameters. We also plan to experiment with dialects of the \ec \ that allow long-term temporal relations between entities and combine \oled \ with weight learning techniques towards online statistical relational learning.

%\vspace*{-0.55cm}

\section*{Acknowledgements}	
This work was partly funded by the EU Project REVEAL (FP7 610928). %We would like to thank the reviewers of the TPLP journal for their valuable comments.
		
\bibliographystyle{acmtrans}
\bibliography{nkatz}

\end{document}